\documentclass[letterpaper]{article} % DO NOT CHANGE THIS
\usepackage{aaai2027}  % DO NOT CHANGE THIS

\usepackage[hyphens]{url}  % DO NOT CHANGE THIS
\usepackage{graphicx} % DO NOT CHANGE THIS
\usepackage{natbib}  % DO NOT CHANGE THIS AND DO NOT ADD ANY OPTIONS TO IT
\usepackage{caption} % DO NOT CHANGE THIS AND DO NOT ADD ANY OPTIONS TO IT
\usepackage{amsmath}

\usepackage{algorithm}
\usepackage{algorithmic}

\usepackage{newfloat}
\usepackage{listings}
\usepackage{booktabs}
\usepackage{multirow}
\usepackage{xcolor}
\usepackage{enumerate}
\usepackage{threeparttable}
\usepackage{cleveref}

\usepackage{pifont}     % \ding{51}/\ding{55} —— 仅 tab_tost.tex 的 ✓/✗ 用
\usepackage{pgfplots}   % 仅 fig_pareto.tex 用
\pgfplotsset{compat=1.18}
\DeclareCaptionStyle{ruled}{labelfont=normalfont,labelsep=colon,strut=off} % DO NOT CHANGE THIS
\floatstyle{ruled}
\newfloat{listing}{tb}{lst}{}
\floatname{listing}{Listing}

\usepackage{booktabs}

\title{Second Thought: Reasoning in Parallel as LLM Agents Act and Observe}
\author{
    Zhensu Sun,
    Chengran Yang,
    Yunbo Lyu,
    Jieke Shi,
    David Lo
}
\affiliations{
    Singapore Management University\\
    80 Stamford Road, Singapore 178902\\
    \{zssun, cryang, yunbolyu, jiekeshi, davidlo\}@smu.edu.sg
}

\begin{document}

\maketitle

\begin{abstract}
LLM agents in the ReAct paradigm alternate between reasoning, acting, and observing, but deliberate reasoning is confined to the Thought phase: while the agent serializes an action and waits for the environment, its reasoning is frozen. We identify this recurring Action–Observation interval as a reasoning idle window and ask whether it can host additional reasoning in parallel that serves future turns. Therefore, we propose Second Thought, a training-free inference framework that forks four auxiliary branches the instant each Thought phase concludes, decodes them concurrently with the main loop, and merges the generated thoughts back when the environment observation arrives. In this way, Second Thought relocates the added reasoning off the main thread's sequential decoding path. Across three agentic benchmarks and three reasoning LLMs, Second Thought lowers the average turn count in all nine model–benchmark pairs and reduces main-thread decoding in six of them by up to 43\% (roughly 20\% on average among those settings), while leaving it essentially unchanged in a seventh; Pass@1 shows no significant change in seven of nine pairs and the two significant
differences are +12.4 and +10.2 points. A paired wall-clock replay confirms that
these reductions can translate into 10.9\% lower median per-task latency. Against a compute-matched control that forces an equivalent budget onto the main thread's own reasoning, it attains strictly higher Pass@1 with 1.3× to 3.2× less sequential decoding in all four settings where the control applies.
\end{abstract}

% Uncomment the following to link to your code, datasets, an extended version or similar.
% You must keep this block between (not within) the abstract and the main body of the paper.
% Make sure that you do not de-anonymize yourself with these links.
\begin{links}
    \link{Code}{https://anonymous.4open.science/r/2nd-thought}
\end{links}

\begin{figure*}[t]
\centering
\includegraphics[width=\linewidth]{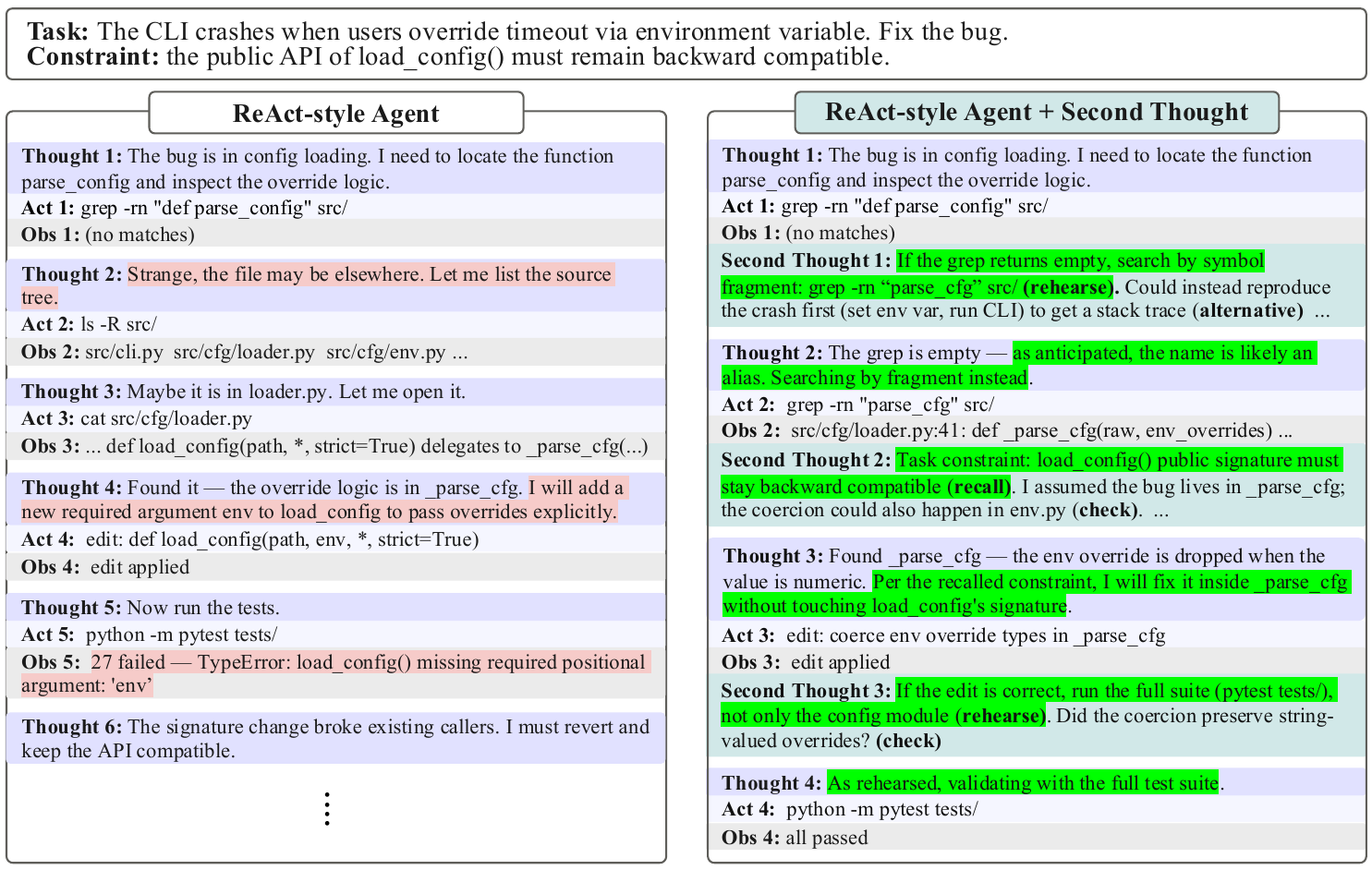}
\caption{An illustrative comparison on a bug-fixing task. \textbf{Left:} a vanilla ReAct-style agent wastes turns on a failed search and breaks backward compatibility (red), requiring a costly revert. \textbf{Right:} with Second Thought, auxiliary reasoning generated during idle windows (green) rehearses fallback searches, recalls the task constraint, and checks assumptions, steering the agent to a correct fix in fewer turns.}
\label{fig:teaser}
\end{figure*}

\section{Introduction}
Large Language Model (LLM)-based agents have emerged as a dominant paradigm for solving complex, multi-step tasks ranging from software engineering to interactive tool use~\cite{lyu2026practitioners}.
Central to their success is the capacity for autonomous reasoning, often structured via text-based thought trajectories that plan, verify, and adapt to a changing environment.
Recent advances demonstrate that scaling inference-time compute yields substantial improvements in problem-solving accuracy.
However, this performance gain comes at a cost: every additional reasoning token is generated on the critical path, so accuracy is purchased with a proportional increase in wall-clock latency.
This prolonged waiting time renders long-reasoning agents slow for interactive use.

Yet an agent is not reasoning throughout its workflow. 
A typical ReAct agent~\citep{yao2023react} operates in a sequential Thought $\rightarrow$ Action $\rightarrow$ Observation loop, and only the Thought phase produces substantive reasoning.
Once a Thought concludes, the agent's thought is naturally frozen: the Action phase materializes the formed plan into a tool invocation, and the Observation phase passively awaits the environment's response.
This creates a recurring \emph{reasoning idle window} on every turn, a period during which additional thinking is known to improve agent performance, yet no thinking takes place at all.
This motivates our question: can this idle window be used for additional parallel reasoning that \emph{complements} the reasoning chain generated so far?

Exploiting this idle window introduces a fundamentally distinct dimension of parallel reasoning that is \emph{orthogonal} to existing paradigms.
Prior techniques, from Self-Consistency~\citep{wang2022self} and Tree-of-Thought search~\citep{yao2023tree} to recent work on population-based selection~\citep{zhou2026opendeepthink} and cross-branch information sharing~\citep{wang2026share}, parallelize within the Thought phase by sampling multiple candidate reasoning chains horizontally.
While effective at expanding the search space, they cannot be directly applied to the reasoning idle window due to its unique nature.
First, because the Thought phase has already concluded and its action is being executed during the idle window, auxiliary reasoning launched here can no longer alter the current turn's decision.
Consequently, rather than competing as alternative branches, it must condition on the established trajectory to serve future turns.
This eliminates the need for complex voting or aggregation steps, allowing auxiliary thoughts to integrate seamlessly via simple concatenation.
Second, the idle window is governed by a hard external deadline—namely, the arrival of the environment observation.
This arrival abruptly terminates any ongoing reasoning, demanding that auxiliary thoughts be \emph{interruption-friendly}: branches cut off mid-generation must still leave behind valid, usable partial results for subsequent turns.

To realize this idea, we propose Second Thought, a training-free inference framework that enables agents to exploit reasoning idle windows for additional reasoning.
Upon the completion of each Thought phase, Second Thought forks four auxiliary reasoning branches that run concurrently with the main loop until the observation returns.
Each branch reasons along a distinct dimension: verifying the assumptions underlying the current plan, rehearsing likely next steps, recalling relevant context from earlier in the trajectory, and drafting contingency plans in case the current one fails.
Unlike horizontal branches, these four branches are not competing solution candidates but complementary perspectives on the same trajectory, so collecting their outputs is mere concatenation.
Figure~\ref{fig:teaser} illustrates the effect: on a bug-fixing task, a vanilla ReAct agent breaks a backward-compatibility constraint and must revert its edit, whereas the recalled constraint and rehearsed fallback produced during idle windows steer the augmented agent directly to a correct fix.
Crucially, every branch is instructed to emit its response as a stream of
\emph{atomic thoughts}: each atomic thought encapsulates a single self-contained concept and depends on no other, so that interrupting a branch mid-generation invalidates at most the thought currently being produced, while all previously completed atomic thoughts remain available.
When the observation arrives for the main loop, the atomic thoughts generated so far are collected and appended to the end of the tool message, allowing the agent to build upon this additional thinking in the next round of reasoning.

In our experiments, we demonstrate the feasibility and value of utilizing the reasoning idle window with Second Thought.
The evaluation is on three agentic benchmarks covering repository-level software engineering (SWE-Bench Pro), terminal operation (Terminal-Bench 2.1), and multi-turn tool-calling dialogue ($\tau^3$-bench), with three reasoning LLMs from distinct model families
(DeepSeek-V4-Flash, Qwen3.6-Plus, and MiniMax-M3).
The results show that the additional reasoning from the idle window offers a better accuracy–latency trade-off than both the unmodified agent and a compute-matched control that places the same or even more reasoning budget on main reasoning.
% Specifically, it reduces the turn count in all nine settings, and reduces main-thread decoding in six while leaving it essentially unchanged in a seventh, by up to 43\% and averaging roughly 20\% among those settings.
% Pass@1 shows no statistically significant change in seven of nine settings and the two significant accuracy gains both occur on Terminal-Bench 2.1 (+12.4 with Qwen3.6-Plus, +10.2 with MiniMax-M3)
% Against compute-matched controls that force an equivalent budget onto the main thread's own reasoning, Second Thought attains strictly higher Pass@1 with $1.3\times$ to $3.2\times$ less sequential decoding, in all four settings where the control is applicable.
Further analysis suggests two distinct sources of benefit: harvested thoughts pre-compute deliberation that would otherwise occupy the next turn's sequential decoding,  while the complementary reasoning dimensions surface considerations the main trajectory does not produce on its own.

Our contributions can be summarized as follows:
\begin{itemize}
    \item We identify and formalize the \emph{reasoning idle window} in ReAct-style
        agent loops, i.e., the recurring Action--Observation interval in which no
        reasoning is generated, and characterize it as unused parallel capacity.
    \item We propose Second Thought, a training-free inference
    framework that utilizes the reasoning idle window through four auxiliary branches (\textsc{Check},
    \textsc{Recall}, \textsc{Rehearse}, and \textsc{Alternative}), whose
        interruption-friendly atomic thoughts are harvested for the next turn, adding reasoning to the trajectory without adding sequential compute to the main thread.
    \item Across three benchmarks and three reasoning LLMs, Second Thought lowers the average turn count in all nine model--benchmark pairs and main-thread sequential
        decoding in seven, with significant Pass@1 gains in two pairs (+12.4 and +10.2, both on Terminal-Bench 2.1) and no significant degradation in other pairs.
\end{itemize}

\section{Related Work}
\subsection{Parallel Reasoning}

Parallel reasoning overcomes the sequential nature of autoregressive decoding by generating multiple reasoning units concurrently.
The dominant paradigm parallelizes \emph{horizontally}: multiple solution paths are explored independently and later aggregated.
Early instantiations sample complete trajectories and merge them by
voting~\citep{wang2022self} or tree
search~\citep{yao2023tree}, while recent work makes branching a native
model capability, allowing the model to spawn and join child threads
for decomposed subtasks~\citep{Sch24,Pan25,Bij25b,Lia25,Yan25,Wu25e}.
Another line relaxes the isolation between branches through shared
attention caches~\citep{Hsu25,Rod25} or a learned organizer--worker
protocol~\citep{Chi25c}.
Across this diversity, the branches remain competing or partitioned
solution attempts whose outcomes must be reconciled by an explicit
merge step, and parallelism serves to shorten the critical path of a
single deliberation phase.
Moreover, parallel reasoning has to wait for the slowest branches, introducing additional latency.
In contrast, the auxiliary branches of Second Thought are not solution
candidates: they extend a single live trajectory from its current
frontier along complementary dimensions, so reintegration is mere
concatenation, with no voting, ranking, or reconciliation.
Closest to our motivation is sleep-time compute~\citep{lin2025sleep0time}, which pre-computes reasoning over a static context before any query arrives, so that test-time latency is reduced for the queries that follow.
Second Thought shares the intuition of relocating deliberation off the critical path, but operates on a window that is intrinsic to the agent loop rather than externally scheduled.

\subsection{Asynchronous and Speculative Agent Execution}
To accelerate the agent loop, recent works propose executing components of the workflow out of order.
\emph{Asynchronous execution} decouples the model from external latency, overlapping token decoding with tool execution~\citep{Gim24,Fen26,Abh24,Xu24,sun2026executing} or reasoning streams with response generation~\citep{Yak25}.
\emph{Speculative execution} anticipates future states, committing predicted actions or reasoning steps only after verification~\citep{Ye25,Pan25d,Hua25b,Zho26b}.
Real-time agent frameworks combine both paradigms, allowing the main reason-and-act thread to continuously issue correctable speculative tool calls while awaiting external updates~\citep{Hoo26}.
Crucially, in all these approaches, the overlapped computation is either the main trajectory's own continuation---tokens that would be generated anyway, simply rescheduled---or a speculative prediction of it, guarded by correction machinery.
Consequently, the total reasoning content along the trajectory remains unchanged, and task accuracy is typically preserved~\citep{Hoo26}.
Second Thought inverts this paradigm: rather than overlapping computation merely to hide tool latency, it leverages this idle time to inject additional reasoning that does not lengthen the critical path, unlike test-time scaling, which places every additional token in sequence
These auxiliary advisory thoughts---which would not otherwise exist---are consumed in the subsequent round rather than committed as actions, thereby requiring no verification or rollback mechanisms.
\begin{figure}[t]
\centering
\includegraphics[width=\linewidth]{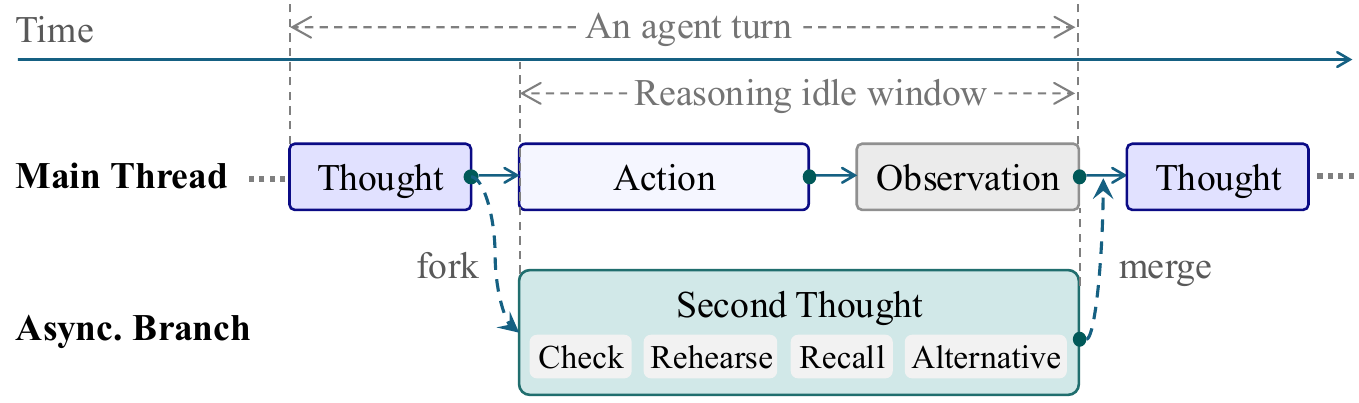}
\caption{Workflow of Second Thought.}
\label{fig:overview}
\end{figure}

\section{Method}
\label{sec:method}

\subsection{Preliminaries: The Reasoning Idle Window}
\label{sec:prelim}
We consider a standard ReAct-style agent~\citep{yao2023react} that solves a task over a sequence of turns. At turn $t$, conditioned on the trajectory history $H_t$, the agent proceeds through three phases:
\begin{itemize}
    \item \textbf{Thought:} the LLM generates explicit reasoning $R_t$ that assesses the current state and plans the next move.
    \item \textbf{Action:} the model serializes an action $A_t$---a structured tool invocation, optionally accompanied by a user-facing description---that instantiates the plan in $R_t$.
    \item \textbf{Observation:} the environment executes $A_t$ and returns feedback $O_t$, which is appended to the history: $H_{t+1} = H_t \cup \{R_t, A_t, O_t\}$.
\end{itemize}
Let $\tau^{\text{think}}_t$, $\tau^{\text{act}}_t$, and $\tau^{\text{obs}}_t$ denote the wall-clock durations of the three phases; the total turn latency is $\tau_t = \tau^{\text{think}}_t + \tau^{\text{act}}_t + \tau^{\text{obs}}_t$.

Crucially, the ReAct format confines deliberate reasoning to the Thought phase. The Action phase merely renders the plan already articulated in $R_t$ into executable form, resolving at most low-level details such as argument literals, while the Observation phase involves no decoding at all---neither produces reasoning directed at subsequent turns. We therefore
refer to this contiguous interval, $W_t = \tau^{\text{act}}_t +
\tau^{\text{obs}}_t$, as the \emph{reasoning idle window}, which recurs on
every turn. Second Thought is designed to capitalize on $W_t$ by
overlapping auxiliary reasoning with the main loop's Action and Observation
phases.

\subsection{Overview}
\label{sec:overview}
Figure~\ref{fig:overview} illustrates the overall workflow of Second Thought. While the main thread advances through the standard Thought--Action--Observation loop, Second Thought runs alongside it on asynchronous auxiliary branches, injecting additional reasoning compute without delaying the critical path. Specifically, the instant the main thread finishes its Thought phase, i.e., at the onset of $W_t$---Second Thought triggers a \emph{fork} operation, spawning four independent branches that each explore a complementary reasoning dimension: \textsc{Check}, \textsc{Recall}, \textsc{Rehearse}, and \textsc{Alternative}. These branches decode concurrently with the main thread's action serialization and tool execution, occupying the entire reasoning idle window; since the window's duration is unpredictable, each branch formats its output as a stream of \emph{atomic thoughts}, self-contained units that remain valid under interruption at any point. Once the observation $O_t$ arrives, a \emph{merge} operation takes place: all active branches are immediately terminated, and the atomic thoughts completed within the window are harvested and appended to the context. Consequently, at turn $t{+}1$, the main thread enters its next Thought phase already in possession of these insights. By construction, Second Thought adds no reasoning tokens to the critical path: branches decode only while the main thread serializes the action and awaits the observation, and the merge reduces to truncating and concatenating text buffers.

\subsection{Forking via Trajectory Continuation}
\label{sec:forking}
Rather than instantiating an external critic to review the trajectory, Second Thought obtains auxiliary reasoning by letting the same model continue its own state. At the fork instant, each branch receives an identical conversation snapshot---the complete history $H_t$ together with the newly generated $R_t$---followed by a short branch-specific instruction. This design carries two practical benefits: the branches inherit the full task context verbatim, requiring no summarization or hand-off protocol; and because all branches share the prompt prefix up to $R_t$ with the main thread, the prefix KV cache is reused across all decoding streams. Finally, branch decoding disables any model-native thinking to emit explicit thoughts as fast as possible before the window closes.

\subsection{Auxiliary Reasoning Protocol}
\label{sec:atoms}
The auxiliary branches must satisfy two critical requirements: (i)~since their execution is strictly bounded by the unpredictable duration of $W_t$, their outputs must remain usable even if interrupted at an arbitrary token; and (ii)~to maximize the utility of the reasoning window, their content should complement, rather than duplicate, the main trajectory's thought process. We address these challenges through \emph{Atomic Thoughts} and \emph{Complementary Reasoning Dimensions}, respectively.

\subsubsection{Atomic Thoughts}
To fulfill the first requirement, Second Thought enforces a unified output contract: every auxiliary branch streams its generation as a sequence of \emph{atomic thoughts}---structurally independent reasoning units, each capturing a single self-contained insight.
To ensure that mid-generation truncation yields valid and usable partial outputs, we enforce two structural constraints.
Firstly, each atomic thought is enclosed in explicit XML tags (\texttt{<thought>}\,\dots\,\texttt{</thought>}), providing unambiguous boundary cues for parsing. Secondly, each thought focuses on a single localized point ($\le 25$ words) and avoids forward or backward references to other units in the stream.
Under this contract, halting a branch mid-generation invalidates at most the single unit currently in flight; all previously closed units remain structurally intact and semantically interpretable. Consequently, the main thread can terminate auxiliary branches unilaterally upon receiving an observation, requiring zero graceful-exit protocols or branch-side coordination.

\subsubsection{Reasoning Dimensions}
\label{sec:dimensions}
Second Thought is inherently modular: the number and semantics of auxiliary branches can be dynamically tailored to specific task characteristics, compute budgets, or domain requirements.
Our reference instantiation for general agentic workflows targets four recurring failure modes of LLM agents, structured along two orthogonal axes: \emph{temporal direction} (retrospective vs.\ prospective) and \emph{scope} (current turn vs.\ overall history). Crossing these axes yields four distinct branches, each mitigating a specific vulnerability:
\begin{itemize}
    \item \textsc{Check} counters \emph{unverified assumptions}, where agents prematurely commit to reasoning that silently presupposes unconfirmed environmental states~\citep{deshpande2025trail0}. It audits the newly finalized $R_t$ for fragile assumptions that the incoming observation might invalidate (e.g., \emph{``assumed the test framework is pytest without checking the configuration''}), flagging potential vulnerabilities without attempting immediate resolution.
    \item \textsc{Recall} counters \emph{context attrition}, where constraints introduced early in a long trajectory lose efficacy as they recede into the context window~\citep{liu2024lost}. It actively resurfaces critical historical context from $H_t$ that remains relevant but may have faded from active attention.
    \item \textsc{Rehearse} counters \emph{unprepared outcomes}, where agents waste extra turns re-planning from scratch when a tool output deviates from expectations~\citep{sun2023adaplanner}. It pre-computes conditional next steps for plausible outcomes (e.g., \emph{``if grep returns no matches, fall back to symbol-based lookup''}), enabling turn $t{+}1$ to execute pre-formulated reactions instantly.
    \item \textsc{Alternative} counters \emph{premature commitment}, where agents stubbornly persist with a failing strategy rather than pivoting~\citep{yao2023tree}. It proactively generates alternative candidate strategies for the active goal along with their trigger conditions, maintaining viable fallback avenues.
\end{itemize}
For specialized deployments, this set can be seamlessly pruned, expanded, or customized (e.g., incorporating a domain-rule verification branch for legal or medical agents).

\subsection{Thought Harvesting}
\label{sec:harvest}
When the observation $O_t$ arrives, Second Thought immediately cancels all in-flight branch generations and harvests their output buffers as they stand. Each buffer is truncated at its last closed \texttt{</thought>} tag, discarding any incomplete unit, and the surviving thoughts are capped at 5 per dimension, bounding per-turn context growth to a small constant. A branch that has not completed a single unit is simply omitted from the harvest; in the limiting case where no branch yields output, the agent degrades to the baseline ReAct loop.
The merged block of atomic thoughts is appended to the end of the tool-observation message. On turn $t{+}1$, the model thus reads its updated history sequentially, i.e., prior reasoning $R_t$, action $A_t$, observation $O_t$, and the attached second thoughts, and conditions its next Thought phase on all of them.

\section{Experiments}
\begin{table}[t]
\centering
\begin{threeparttable}
\setlength{\tabcolsep}{2pt}
\caption{Overall performance of Second Thought across three benchmarks. $\text{\#OUT}_{\text{main}}$ and \#Turns respectively indicate the number of output tokens and turns of the main thread. Cells marked ``---'' denote settings where the $s1$ control is not applicable. We use Benjamini–Hochberg corrected McNemar test for Pass@1 and Wilcoxon signed-rank for rest metrics.}
\label{tab:main}
\begin{tabular}{llcccc}
\toprule
\textbf{Bench} & \textbf{Model} & \textbf{Setting} & \textbf{Pass@1} & \textbf{$\text{\#OUT}_{\text{main}}$} & \textbf{\#Turns} \\
\midrule
\multirow{9}{*}{SWE-Pro} & \multirow{3}{*}{DS-V4} & base & 48.7 & 23{,}841 & 56.2 \\
 &  & s1 & 49.3 & 54{,}463* & 58.5 \\
 &  & ours & \textbf{52.0} & \textbf{20{,}255}* & \textbf{52.8}* \\
\cmidrule(lr){2-6}
 & \multirow{3}{*}{Qwen3.6} & base & \textbf{52.0} & 36{,}519 & 57.1 \\
 &  & s1 & 48.7 & 65{,}634* & 55.7 \\
 &  & ours & 51.3 & \textbf{20{,}798}* & \textbf{50.6}* \\
\cmidrule(lr){2-6}
 & \multirow{3}{*}{MM-M3} & base & \textbf{55.3} & 15{,}400 & 79.6 \\
 &  & s1 & --- & --- & --- \\
 &  & ours & \textbf{55.3} & \textbf{14{,}579} & \textbf{71.9}* \\
\midrule
\multirow{9}{*}{TB2} & \multirow{3}{*}{DS-V4} & base & 50.6 & 48,202 & 40.2 \\
 &  & s1 & 50.6 & 68{,}003* & \textbf{33.4}* \\
 &  & ours & \textbf{52.8} & \textbf{32{,}892}* & 35.8* \\
\cmidrule(lr){2-6}
 & \multirow{3}{*}{Qwen3.6} & base & 39.3 & \textbf{25{,}158} & 25.5 \\
 &  & s1 & 46.1 & 40{,}642* & \textbf{22.1} \\
 &  & ours & \textbf{51.7}* & 31{,}396 & 24.0 \\
\cmidrule(lr){2-6}
 & \multirow{3}{*}{MM-M3} & base & 49.4 & \textbf{36{,}686} & 44.6 \\
 &  & s1 & --- & --- & --- \\
 &  & ours & \textbf{59.6}* & 36{,}705 & \textbf{43.1} \\
\midrule
\multirow{6}{*}{$\tau^3$-bench} & \multirow{2}{*}{DS-V4} & base & \textbf{24.0} & 16{,}260 & 27.5 \\
 &  & ours & \textbf{24.0} & \textbf{15{,}203} & \textbf{26.3} \\
\cmidrule(lr){2-6} 
 & \multirow{2}{*}{Qwen3.6} & base & 16.7 & 16{,}755 & 26.1 \\
 &  & ours & \textbf{19.8} & \textbf{13{,}764}* & \textbf{25.9} \\
\cmidrule(lr){2-6}
 & \multirow{2}{*}{MM-M3} & base & 17.7 & \textbf{10{,}004} & 24.0 \\
 &  & ours & \textbf{20.8} & 10{,}367 & \textbf{20.0}* \\
\bottomrule
\end{tabular}
\begin{tablenotes}
\small
\item $^* p < 0.05$
\end{tablenotes}
\end{threeparttable}
\end{table}

\subsection{Experimental Setup}
\subsubsection{Benchmarks.}
We evaluate on three agentic benchmarks chosen to cover different tasks.
\emph{SWE-Bench-Pro}~\citep{deng2025swe} is repository-level software
engineering: each instance places the agent in a per-instance Docker
container, where it must localize and patch a real bug, graded by held-out
fail-to-pass tests. We use a randomly sampled subset of 150 instances with a
100-turn step limit. 
\emph{Terminal-Bench 2.1}~\citep{merrill2026terminal} comprises 89 containerized
terminal-operation tasks, such as system administration, data processing, and software builds, graded by task-specific verification scripts.
\emph{$\tau^3$-bench}~\citep{barres2025tau} (banking domain, 97 tasks) is multi-turn customer-service dialogue grounded in knowledge retrieval: the agent converses with an LLM-simulated user while issuing function calls, and must ground its responses in policy documents retrieved from an unstructured corpus.

\subsubsection{Agent.}
Each agent setup pairs a reasoning LLM with a benchmark-specific harness. We evaluate three reasoning LLMs from distinct model families: DeepSeek-V4-Flash~\citep{deepseek-ai2026deepseek0v40}, Qwen3.6-Plus~\citep{qwen36plus}, and MiniMax-M3~\cite{minimax2026minimaxm3}, all accessed via streaming APIs with their native reasoning/thinking modes enabled. 
For the execution harnesses, SWE-Bench-Pro employs mini-SWE-agent~\citep{yang2024sweagent}, Terminal-Bench 2.1 uses its official execution harness, and $\tau^3$-bench uses its standard function-calling dialogue loop paired with the benchmark's user simulator.

\subsubsection{Baselines.}
We compare against two baselines. \emph{base} is the unmodified agent.
\emph{s1} adapts budget forcing~\citep{muennighoff2025s1} to serve as a
compute-matched control: it spends the same additional reasoning budget
as Second Thought, but places it on the critical path as a continuation
of the turn's own thought.
Concretely, whenever the main Thought would terminate before consuming the token budget that Second Thought's branches produce in that turn, we suppress the end-of-thinking delimiter and force the model to continue reasoning until its thought length reaches at least that budget.
This baseline isolates the effect of the amount of reasoning tokens.
Notably, budget forcing requires suppressing the end-of-thinking delimiter and having the model continue its own thought from a prefilled prefix, but MiniMax-M3 does not honor an unclosed-thought prefix and emits a final answer instead of extending its deliberation, while on $\tau^3$-bench prefix continuation cannot be combined with function-calling requests.
Therefore, \emph{s1} is not available in these two settings.

\subsection{Main Results}
Table~\ref{tab:main} presents the overall performance of Second Thought against the standard ReAct baseline and the compute-matched reasoning-extended baseline ($s1$) across nine benchmark--model combinations. 
Second Thought reduces the average turn count in all nine pairs, and reduces main-thread decoding in six of them by up to 43\% (from 36.5k to 20.8k tokens on SWE-Bench Pro with Qwen3.6-Plus) and by roughly 20\% on average among those settings. Pass@1 is preserved or improved in eight; the single decrease (from 52.0\% to 51.3\% on SWE-Bench Pro with Qwen3.6-Plus) amounts to one instance out of 150 and is not statistically significant, whereas the largest gain (+12.4 points on Terminal-Bench 2.1 with Qwen3.6-Plus) is. Notably, the only substantive increase in main-thread decoding (+24.8\%) occurs in exactly that setting, which suggests the harvested thoughts do not merely displace tokens the main thread would have produced, but can also open lines of reasoning it would not have pursued at all. 
Depending on how scaling reasoning length affects task accuracy, the overall trends fall into three distinct regimes.

On tasks where simply extending the reasoning trace length ($s1$) hurts accuracy or inflates the main thread's own decoding (e.g., SWE-Bench Pro with Qwen3.6-Plus, where $s1$ drives output tokens from 36,519 to 65,634 yet drops Pass@1 from $52.0\%$ to $48.7\%$), Second Thought effectively suppresses redundant outputs while preserving accuracy.
On Terminal-Bench 2.1 with DeepSeek-V4-Flash, it boosts Pass@1 to $52.8\%$ ($+2.2\%$ over both base and $s1$) while reducing main-thread output tokens to 32,892, $32\%$ below the baseline agent and $2.1\times$ fewer than $s1$.
On tasks where extended reasoning improves accuracy (e.g., Terminal-Bench 2.1 with Qwen3.6-Plus), Second Thought delivers even superior Pass@1 gains ($+12.4$ points, $+5.6$ over $s1$) while decoding $1.3\times$ fewer main-thread tokens than $s1$ (31,396 vs.\ 40,642) and taking fewer turns than the baseline agent (24.0 vs.\ 25.5).
On $\tau^3$-bench (banking), Second Thought achieves smaller Pass@1 gains and improves it by up to $+3.1\%$ (Qwen3.6-Plus and MiniMax-M3) at comparable or lower main-thread token counts and consistently fewer turns.
We attribute this asymmetry to the nature of the domain: this domain has short idle window and banking failures stem largely from retrieval quality and policy adherence rather than from the planning errors that \textsc{Check}, \textsc{Rehearse}, and \textsc{Alternative} target, so the harvested thoughts shorten deliberation without substantially changing which documents are retrieved.
This points to a promising future direction: dynamically selecting branch types based on the domain.

\begin{figure}[t]
\centering
\includegraphics[width=\linewidth]{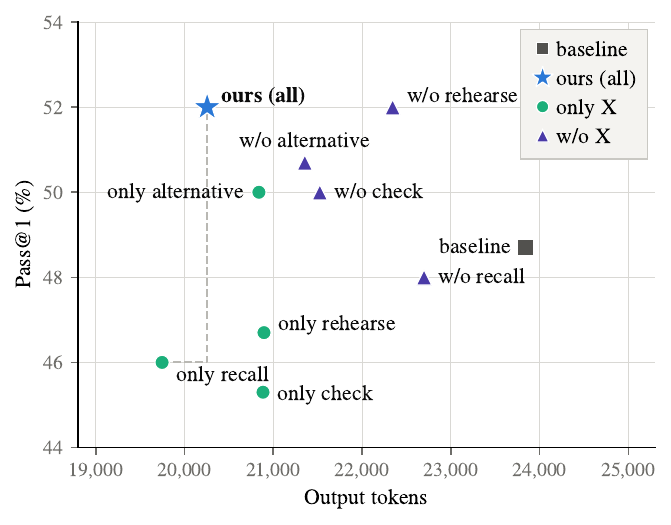}
\caption{Single-dimension (only-X) and leave-one-out (w/o-X) variants on SWE-Bench Pro with DeepSeek-V4-Flash; upper-left is better}
\label{fig:ablation}
\end{figure}

\subsection{Ablation Study}
Figure~\ref{fig:ablation} evaluates the trade-offs of single-dimension (\emph{only-X}) and leave-one-out (\emph{w/o-X}) variants using DeepSeek-V4-Flash on SWE-Bench Pro.
Restricting to a single branch generally fails to match the full configuration's performance, with most single-dimension variants falling near or below the baseline ($48.7\%$ Pass@1).
Specifically, \emph{only-recall} attains the fewest main-thread output tokens ($19.7$k) but drops Pass@1 to $46.0\%$, indicating that passive context retrieval without active validation or contingency planning is insufficient for accurate action selection. Conversely, leave-one-out experiments highlight the complementary roles of each component: removing \textsc{Recall} (\emph{w/o-recall}) causes the sharpest accuracy drop ($48.0\%$ Pass@1), suggesting that resurfacing historical constraints helps prevent repeated errors, whereas removing \textsc{Rehearse} (\emph{w/o-rehearse}) maintains Pass@1 ($52.0\%$) but inflates main-thread output from $20.3$k to $22.4$k tokens ($+10\%$), confirming that pre-computing conditional next steps primarily offloads deliberation the main thread would otherwise perform turn by turn.
Overall, the full configuration is Pareto-optimal: no variant attains higher Pass@1, and the only variant that decodes fewer main-thread tokens (\emph{only-recall}, $19.7$k) sacrifices $6.0$ points of accuracy for a $2.5\%$ saving.

\subsection{In-depth Analysis}

\subsubsection{Measured Timing Analysis.}
Raw wall-clock time from our main results is not comparable across our runs, which span different time periods with drifting API serving speed and fluctuating execution load on a shared server.
We therefore replay our experiment with a controlled setting to see if the reduced output tokens and turns can be translated to real reduction in latency.
We replay $50$ SWE-Bench Pro instances with DeepSeek-V4-Flash under a paired protocol: for each instance, \emph{base} and \emph{ours} run back-to-back within the same window against the same endpoint at a fixed concurrency of $4$, repeated $3$ times, and we report the median so that residual drift affects both settings alike.
Under this protocol, Second Thought reduces median per-task wall-clock time from $256.9$\,s to $229.0$\,s ($-10.9\%$), decomposing along exactly the two quantities of Table~\ref{tab:main}: main-thread decoding time drops from $168.7$\,s to $146.1$\,s ($-13.4\%$), broadly in proportion to the $15.0\%$ fewer main-thread output tokens, while tool execution time drops from $71.6$\,s to $67.3$\,s ($-6.0\%$), tracking the decrease from $56.2$ to $52.8$ turns at an unchanged $1.27$\,s per turn.
The residual gap between the $15.0\%$ token reduction and the $13.4\%$ time reduction is contention: running four branches lowers main-thread decoding throughput from $141.3$ to $138.6$ tokens/s ($-1.9\%$), costing $2.8$\,s against the $27.9$\,s saved per task.
This modest penalty can be attributed to branches decoding with native thinking disabled and sharing the prompt prefix KV cache with the main thread, so their serving load is dominated by cached prefill rather than sustained decoding.
Overall, this demonstrates that the savings reported in Table~\ref{tab:main} are not an accounting artifact of where tokens are attributed, but a genuine shortening of the critical path.

\subsubsection{Reasoning Substitution Effect.}
We measure the per-turn effect of harvested second thoughts with a replay experiment.
We randomly sample 100 trajectories generated by DeepSeek-V4-Flash on SWE-Bench Pro.
For each trajectory, we replay the turn that follows a harvest under two conditions: with the harvested thoughts in the context, or without them.
The experiment is repeated three times.
Removing these thoughts inflates next-turn reasoning volume from $196.2$ to $316.5$ tokens on average.
This suggests that second thoughts substitute for reasoning that the agent would otherwise generate on the critical path of the next turn.

\begin{table}[t]
\centering
\setlength{\tabcolsep}{1.5pt}
\caption{Controlled comparison on SWE-Bench Pro with DeepSeek-V4-Flash. w/o $R_t$: branches see $H_t$ but not the just-finalized
$R_t$. fork onset: fork at turn start, widening the time window to cover the Thought phase.
unbounded: the main thread stalls until
branches close, so branch tokens fall on the critical path.}
\label{tab:fork-conditioning}
\newcommand{\hdr}[1]{\multirow{2}{*}[-2pt]{\textbf{#1}}}
\begin{tabular}{lccccc}
\toprule
\hdr{Method} & \hdr{Pass@1} & \multicolumn{2}{c}{\textbf{\#Output tokens}} & \hdr{\#Turns} & \hdr{\#Atoms} \\
\cmidrule(lr){3-4}
& & \textbf{Main} & \textbf{Branch} & & \\
\midrule
baseline & 48.7 & 23.8k & --- & 56.2 & 0 \\
s1 & 49.3 & 54.5k & --- & 58.5 & 0 \\
ours & 52.0 & \textbf{20.3k} & 13.6k & 52.8 & 376 \\
\midrule
- w/o $R_t$ & 49.3 & 23.9k & 21.4k & 52.6 & 607 \\
- fork onset & 51.3 & 21.8k & 48.6k & 47.3 & 1{,}418 \\
- unbounded & \textbf{56.7} & 48.4k & --- & \textbf{43.6} & 846 \\
\bottomrule
\end{tabular}
\end{table}
\begin{table}[t]
\centering
\small
\setlength{\tabcolsep}{1pt}
\caption{Per-task API cost of Second Thought on SWE-Bench Pro, computed over the same runs as Table~\ref{tab:main} under each provider's token prices on 1 Jul 2026.}
\label{tab:cost}
\begin{tabular}{lcccl}
\toprule
\multirow{2}{*}{\textbf{Setting}} & \multicolumn{3}{c}{\textbf{\#Tokens per task}} & \multirow{2}{*}{\textbf{Cost per task}} \\
\cmidrule(lr){2-4}
 & Uncached & Cached & Generated & \\
\midrule
\multicolumn{5}{l}{\textit{DeepSeek-V4-Flash} \;\;\scriptsize \$0.140 / \$0.0028 / \$0.280 per M token} \\
base                   & 32.5k & 1.27M & 23.8k & \$0.015 \\
ours (4 branches) & 59.0k & 2.45M & 33.8k & \$0.025 ($+66.4\%$) \\
ours (1 branch)  & 40.9k & 1.60M & 25.0k & \$0.017 ($+16.3\%$) \\
\midrule
\multicolumn{5}{l}{\textit{Qwen3.6-Plus} \;\;\scriptsize \$0.325 / \$0.325 / \$1.950 per M token} \\
base          & 34.0k & 1.33M & 36.5k & \$0.514 \\
ours (4 branches) & 92.3k & 3.70M & 32.4k & \$1.297 ($+152.2\%$) \\
ours (1 branch) & 50.3k & 1.82M & 23.7k & \$0.653 ($+27.0\%$) \\
\midrule
\multicolumn{5}{l}{\textit{MiniMax-M3} \;\;\scriptsize \$0.300 / \$0.060 / \$1.200 per M token} \\
base                   & 43.3k & 2.13M & 15.4k & \$0.159 \\
ours (4 branches) & 86.6k & 6.49M & 27.1k & \$0.448 ($+181.5\%$) \\
ours (1 branch)  & 52.5k & 2.97M & 18.0k & \$0.216 ($+35.5\%$) \\
\bottomrule
\end{tabular}
\end{table}

\subsubsection{Idle Window Utilization.}
We characterize the operational behavior of Second Thought across the three benchmarks and agents.
Harvest yield directly scales with the duration of the reasoning idle window; when turns are binned by window length relative to their run median, the fraction yielding a non-empty harvest rises monotonically from $0.0\%$ to $90.1\%$.
One fixed configuration thus covers windows differing by orders of magnitude across three model families and three benchmarks, with no per-cell tuning.
Coverage accordingly follows the idle time rather than the turn count: the $28.7\%$ of turns that harvest hold $86.7\%$ of it, and $96.5\%$ of tasks receive second thoughts at least once.

\subsubsection{Disentangling Window Duration  from Fork Conditioning.}

Second Thought couples three choices that jointly determine what the branches produce: when they fork, what context they receive at the fork instant, and whether the arriving observation truncates them. Two of these vary the same underlying quantity -- how much time the branches are given -- by different means, and Table~\ref{tab:fork-conditioning} separates it from
the context they receive.
Forking at turn onset buys a wider window by
covering the Thought phase, but pays for it: the branches no longer see $R_t$.
Isolating that penalty at a fixed window, stripping $R_t$ costs 2.7 points
(52.0\% to 49.3\%) even though it yields 1.6x the atoms (11.5 vs 7.1 per
turn), since a branch writing without sight of the plan about to execute
produces more units of less use. Forking early incurs exactly this penalty
yet still reaches 51.3\%, within one instance of our default: the wider
window is worth roughly +2.0 and nearly cancels it. Removing truncation
isolates the window effect at fixed conditioning, and it achieves 56.7\% Pass@1, +4.7 over our
default.
Window duration is therefore the dominant lever, and our default is
window-starved, capturing 41\% of the attainable gain over baseline (+3.3 of
+8.0). The remainder is priced rather than unreachable: holding the main
thread until the branches close moves their entire budget onto the critical
path, raising sequential decoding from 20.3k to at most 48.4k tokens. Second
Thought instead takes the endpoint at which that cost is zero, and the gap
closes wherever the environment affords longer action--observation
intervals.

\subsubsection{Cost Analysis.}
\label{sec:cost}
Table~\ref{tab:cost} summarizes per-task API costs on SWE-Bench Pro.
Running four auxiliary branches increases the total API cost by 66.4\% to 181.5\% across models. This increase is almost entirely driven by input prompt processing rather than output generation.
In fact, output token costs vary by less than \$0.02 per task across all models.
Instead, the cost overhead is dominated by cached prefix reads and tracks the provider's cache discount heavily.
Since all four branches share the main thread's prefix KV cache, these reads incur near-zero marginal compute.
For resource-constrained settings, keeping only the top-performing branch (ALTERNATIVE) cuts the added cost down to 16.3\%–35.5\%.

\section{Conclusion}
We identify the reasoning idle window, i.e., the recurring action–observation interval within ReAct-style loops during which no reasoning is produced.
We then propose Second Thought, a training-free framework that populates this window with four auxiliary branches: Check, Recall, Rehearse, and Alternative. Interruption-safe atomic thoughts generated by these branches are collected and reused for the subsequent reasoning turn.
Evaluated across three benchmarks and three reasoning LLMs, Second Thought
lowers turn count in all nine configurations (significant in five) and
main-thread decoding in six (significant in four), while leaving Pass@1
statistically unchanged in seven of nine and significantly improved in two.

\bibliography{aaai2027}

% Check whether the conference requires a reproducibility checklist to be included in the paper.
% If so, you can uncomment the following line and ajust the path to include it.
% \input{ReproducibilityChecklist.tex}

\end{document}